\documentclass{article}

\usepackage{microtype}
\usepackage{graphicx}
\usepackage{subfigure}
\usepackage{amsfonts}
\usepackage{amsmath}
\usepackage{booktabs} 

\usepackage{hyperref}

\usepackage[accepted]{icml2020}

\icmltitlerunning{Doubly Stochastic Variational Inference for Neural Processes with Hierarchical Latent Variables}

\begin{document}

\twocolumn[
\icmltitle{Doubly Stochastic Variational Inference for Neural Processes \\ with Hierarchical Latent Variables}



\icmlsetsymbol{equal}{*}

\begin{icmlauthorlist}
\icmlauthor{Qi Wang}{to}
\icmlauthor{Herke van Hoof}{to}
\end{icmlauthorlist}

\icmlaffiliation{to}{Amsterdam Machine Learning Lab, University of Amsterdam, Amsterdam, the Netherlands}

\icmlcorrespondingauthor{Qi Wang}{q.wang3@uva.nl}

\icmlkeywords{Machine Learning, ICML}

\vskip 0.3in
]



\printAffiliationsAndNotice{} 

\begin{abstract}
Neural processes (NPs) constitute a family of variational approximate models for stochastic processes with promising properties in computational efficiency and uncertainty quantification. These processes use neural networks with latent variable inputs to induce predictive distributions.
However, the expressiveness of vanilla NPs is limited as they only use a global latent variable, while target-specific local variation may be crucial sometimes. To address this challenge, we investigate NPs systematically and present a new variant of NP model that we call Doubly Stochastic Variational Neural Process (DSVNP). This model combines the global latent variable and local latent variables for prediction. We evaluate this model in several experiments, and our results demonstrate competitive prediction performance in multi-output regression and uncertainty estimation in classification.
\end{abstract}

\section{Introduction}
 In recent decades, increasingly attention has been focused on deep neural networks, and the success of deep learning in computer vision, natural language processing and robotics control etc. can be attributed to the great potential of function approximation with high-capacity models \cite{lecun2015deep}. Despite this, there still remain some limitations which incur doubts from industry when applying these models to real world scenarios. Among them, uncertainty quantification is long-standing and challenging, and instead of point estimates we prefer probabilistic estimates with meaningful confidence values in predictions. With uncertainty estimates at hand, we can relieve some risk and make relatively conservative choices in cost-sensitive decision-making \cite{gal2016dropout}. 

Faced with such reality, Bayesian statistics provides a plausible schema to reason about subjective uncertainty and stochasticity, and marrying deep neural networks and Bayesian approaches together satisfies practical demands. Traditionally, Gaussian processes (GPs) \cite{rasmussen2003gaussian} as typical non-parametric models can be used to handle uncertainties by placing Gaussian priors over functions. The advantage of introducing distributions over functions lies in the characterization of the underlying uncertainties from observations, enabling more reliable and flexible decision-making. For example, the uncertainty-aware dynamics model enjoys popularity in model-based reinforcement learning, and GPs deployed in PILCO enable propagation of uncertainty in forecasting future states \cite{deisenroth2011pilco}. Another specific instance can be found in demonstration learning; higher uncertainty in prediction would suggest the learning system to query new observations to avoid dangerous behaviors \cite{thakur2019uncertainty}. During the past few years, a variety of models inspired by GPs and deep neural networks have been proposed  \cite{salimbeni2017doubly,snelson2006sparse,titsias2009variational,titsias2010bayesian}. 

However, GP induced predictive distributions are met with some concerns. One is high computational complexity in prediction due to the matrix inversion, and another is less flexibility in function space. Recognized as an explicit stochastic process model, the vanilla GP strongly depends on the assumption that the joint distribution is Gaussian, and such a unimodal property makes it tough to scale to more complicated cases. These issues facilitate the birth of adaptations or approximate variants for GP related models  \cite{garnelo2018conditional,garnelo2018neural,louizos2019functional}, which incorporate latent variables in modeling to account for uncertainties. Among them, Neural Processes (NPs) are relatively representative with advantages like uncertainty-aware prediction and efficient computations. 

In this paper, we investigate NP related models and explore more expressive approximations towards general stochastic processes ($\mathcal{SP}$s). \textit{The main focus is on a novel variational approximate model for NPs to solve learning problems in high-dimensional cases.} To improve flexibility in predictive distributions, hierarchical latent variables are considered as part of the model structure. 
Our primary contributions can be summarized as follows.
\begin{itemize}
    \item We systematically revisit NPs, $\mathcal{SP}$s and other related models from a unified perspective with an implicit Latent Variable Model (LVM). Both GPs and NPs are studied in this hierarchical LVM.
    \item The Doubly Stochastic Variational Neural Process (DSVNP) is proposed to enrich the NP family, where both global and target specific local latent variables are involved in a predictive distribution. 
    \item Experimental results demonstrate effectiveness of the proposed Bayesian model in high dimensional domains, including regression with multiple outputs and uncertainty-aware image classification.
\end{itemize}

\section{Deep Latent Variable Model as Stochastic Processes}

\begin{figure*}[ht]
\vskip 0.1in
\begin{center}
\centerline{\includegraphics[width=0.80\textwidth,height=0.21\textheight]{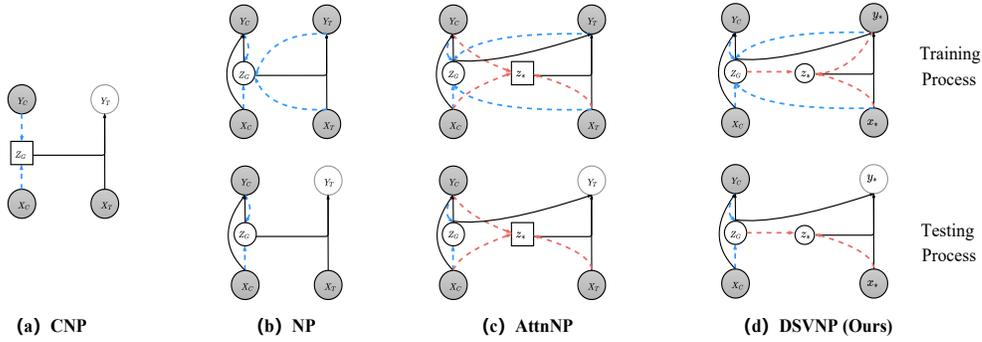}}
\caption{Probabilistic Graphs for CNP, vanilla NP, Attentive NP and DSVNP. The blue dotted lines characterize the inference towards global latent variable $z_G$, while the pink dotted lines are for target specific local latent variables $z_*$. The ones in first row are training cases, while those in second row are testing cases.}
\label{fig1}
\end{center}
\vskip -0.3in
\end{figure*}

Generally, a stochastic process places a distribution over functions and any finite collections of variables can be associated with an implicit probability distribution.
Here, we naturally formulate an implicit LVM \cite{rezende2014stochastic,kingma2014stochastic} to characterize General Stochastic Function Processes (GSFPs). The conceptual generation paradigm for this LVM can be depicted in the following equations,
\begin{equation}\label{eq_3}
z_i=\phi(x_i)+\epsilon(x_i)
\end{equation}
\begin{equation}\label{eq_5}
y_i=\varphi(x_i,z_i)+\zeta_i
\end{equation}
where terms $\epsilon$ and $\zeta$ respectively indicate the stochastic component in the latent space and random noise in observations. To avoid ambiguity in notation, the stochastic term $\epsilon$ is declared as an index dependent random variable $\epsilon(x_i)$, and $\zeta_i$ is observation noise in the environment. Also, the transformations $\phi$ and $\varphi$ are assumed to be Borel measurable, and the latent variables in Eq. (\ref{eq_3}) are not restricted. \textit{Note that they can be some set of random variables with statistical correlations without loss of generality.} When Kolmogorov Extension Theorem \cite{oksendal2013stochastic} is satisfied for ${\epsilon(x_i)}$, a latent $\mathcal{SP}$ can be induced. Eq. (\ref{eq_3}) decomposes the process into a deterministic component and a stochastic component in some models. The transformation $\varphi$ in Eq. (\ref{eq_5}) is directly connected to the output. 
Such a generative process can simultaneously inject \textit{aleatoric uncertainty} and \textit{epistemic uncertainty} in modelling \cite{hofer2002approximate}, but inherent correlations in examples make the exact inference intractable mostly.

Another principal issue is about prediction with permutation invariance, which learns a conditional distribution in $\mathcal{SP}$ models. With the context $\mathcal{C}=\{(x_i,y_i)\vert i=1,2,\dots,N\}$ and input variables of the target $x_T$, we seek a stochastic function $f_{\theta}$ mapping from $X$ to $Y$ and formalize the distribution as $p_{\theta}(y_T\vert x_C,y_C,x_T)$\footnote{For brief notations, the inputs, outputs of the context and the target are respectively denoted as $x_C=x_{1:N}$, $y_C=y_{1:N}$, $x_T=x_{1:N+M}$, $y_T=y_{1:N+M}$. Only in CNP, $x_T=x_{N+1:N+M}$, $y_T=y_{N+1:N+M}$. And $[x_*,y_*]$ refers to any instance in the target.} invariant to the order of context observations. The definitions about permutation invariant functions (PIFs) and permutation equivariant functions (PEFs) are included in Appendix A.

\subsection{Gaussian Processes in the Implicit LVM}
Let us consider a trivial case in the LVM when the operation $\varphi$ is an identity map, $\zeta$ is Gaussian white noise, and the latent layer follows a multivariate Gaussian distribution. This degenerated case indicates a GP, and terms $\phi$, $\epsilon$ are respectively the mean function and the zero-mean GP prior. Meanwhile, recall that the prediction at target input $x_T$ in GPs relies on a predictive distribution $p(y_T|x_C,y_C,x_T)$, where the mean and covariance matrix are inferred from the context $[x_C,y_C]$ and target input $x_T$.
\begin{equation}\label{eq_7}
\begin{split}
\mu(x_T;x_C,y_C)=\phi_{\theta}(x_T)+\Sigma_{T,C}\Sigma_{C,C}^{-1}\big(y_C-\phi_{\theta}(x_C)\big) \\
\Sigma(x_T;x_C,y_C)=\Sigma_{T,T}-\Sigma_{T,C}\Sigma_{C,C}^{-1}\Sigma_{C,T}  
\end{split}
\end{equation}
Here $\phi_{\theta}$ and $\Sigma$ in Eq. (\ref{eq_7}) are vectors of mean functions and covariance matrices. For additive terms, they embed context statistics and connect them to the target sample $x_T$. Furthermore, two propositions are drawn, which we prove in Appendix B.

\textbf{Proposition 1.} The statistics of GP predictive distributions, such as mean and (co)-variances, for a specific point $x_*$ are PIFs, while those in $p(y_T|x_C,y_C,x_T)$ are PEFs.

\subsection{Neural Processes in the Implicit LVM}
In non-GP scenarios, inference and prediction processes for the LVM can be non-trivial, and NPs are the family of approximate models for implicit $\mathcal{SP}$s. Also, relationship between GPs and NPs can be explicitly established with deep kernel network \cite{rudner2018connection}. Note that \textit{NPs translate some properties of GPs to predictive distributions, especially permutation invariance of context statistics}, which is highlighted in \textbf{Proposition 1}. Here three typical models are investigated, respectively conditional neural process (CNP) \cite{garnelo2018conditional},vanilla NP \cite{garnelo2018neural} and attentive neural process (AttnNP) \cite{DBLP:conf/iclr/KimMSGERVT19}.

When approximate inference is used in NP family with Latent Variables, a preliminary evidence lower bound (ELBO) for the training process can be derived, which aims at predictive distributions for most NP related models.
\begin{equation}\label{eq_6}
\begin{split}
\ln\big[p(y_T\vert x_C,y_C,x_T)\big]\geq \mathbb{E}_{q_{\phi}(z_T)}\ln\big[p_{\theta}(y_T\vert x_T,z_T)\big] \\
-D_{KL}\big[q_{\phi}(z_T\vert x_C,y_C,x_T,y_T)\parallel p(z_T\vert x_C,y_C,x_T)\big]
\end{split}
\end{equation}

To ensure context information invariant to orders of points, CNP embeds the context point $[x_C,y_C]$ in an elementwise way and then aggregates them with a permutation invariant operation $\oplus$, such as mean or max pooling.
\begin{equation}\label{eq_8}
r_i=h_{\theta}(x_i,y_i),\quad r_C=\bigoplus_{i=1}^{N}r_i
\end{equation}
The latent variable in CNP is a deterministic embedding of the form $p_{\theta}(z_C\vert x_C,y_C)=r_C(x_C,y_C)$. Followed with Eq. (\ref{eq_5}), CNP decodes statistics as the mean and the variance for a predictive distribution.

For vanilla NPs, the encoder structure resembles that of CNPs, and the learned embedding variable $r$ in Eq. (\ref{eq_8}) is no longer a function but a Gaussian variable after amortized transformations. For the graphical structure of this LVM in vanilla NPs (Refer to Figure \ref{fig1} (b)), all latent variables are degraded to a global Gaussian latent variable, which accounts for the consistency. 

AttnNP further improves the expressiveness of context information in NP, leaving the latent variable as the combination of a global variable and a local variable. 
Especially, the attention network uses self-attention or dot-product attention to enable transformations of context points and the extraction of hierarchical dependencies between context points and target points. For the graphical model of this LVM in AttnNP, the context information is instance-specific. The latent variable of AttnNP in Eq. (\ref{eq_9}) is the concatenation of attention embedding $z_{attn}$ from element-wise context embedding $s_i$ and a global latent variable $z_{G}$ drawn from an amortized distribution parameterized with Eq. (\ref{eq_8}). 
\begin{equation}\label{eq_9}
    z_{attn}=\bigoplus_{i=1}^{N}w(x_i,x_*)s_i,\quad z=[z_{attn},z_{G}]
\end{equation}
As a summary, AttnNP boosts performance with attention networks, which implicitly seeks more flexible functional translations for each target.

\begin{table*}[t]
\caption{Structure Summary over NP Related Models on Training Dataset. Here $f$ corresponds to some functional operation. Global latent variable in CNP only governs points to predict, while that in NP works for the whole points. And local latent variables in AttnNP and DSVNP are distinguished, with the latter as a latent random variable.}
\label{np_table}
\vskip 0.15in
\begin{center}
\begin{small}
\begin{sc}
\begin{tabular}{lllll}
\toprule
NP Family & Recognition Model & Generative Model & Prior Distribution & Latent Variable \\
\midrule
CNP& $z_C=f(x_C,y_C)$& $p(y_T\vert z_C,x_T)$& NULL& Global \\
NP& $q(z_G\vert x_C,y_C,x_T,y_T)$& $p(y_T\vert z_G,x_T)$& $p(z_G\vert x_C,y_C)$& Global\\
AttnNP& $q(z_G\vert x_C,y_C,x_T,y_T)$, & $p(y_*\vert z_G,z_*,x_*)$& $p(z_G\vert x_C,y_C)$&  Global \\
 &$z_*=f(x_C,y_C,x_*)$ & & &+Local \\
DSVNP (Ours)& $q(z_G\vert x_C,y_C,x_T,y_T)$, & $p(y_*\vert z_G,z_*,x_*)$& $p(z_G\vert x_C,y_C)$, & Global\\
 &$q(z_*\vert z_G,x_*,y_*)$ & &$p(z_*\vert z_G,x_*)$ &+Local \\
\bottomrule
\end{tabular}
\end{sc}
\end{small}
\end{center}
\vskip -0.1in
\end{table*}

\subsection{Connection to Other Models}
In some scenarios, when the latent layer in Eq. (\ref{eq_3}) is specified as a Markovian chain, the LVM degrades to classical state space model. If random variables in the latent layer of the LVM are independent, the resulted neural network is similar to the conditional variational auto-encoder \cite{sohn2015learning} and no context information is utilized for prediction. Instead, the existence of correlations between latent variables in the hidden layer increases the model capacity. 
The induced $\mathcal{SP}$ in Eq. (\ref{eq_5}) is a warped GP when the latent $\mathcal{SP}$ is a GP and the transformation $\varphi$ is nonlinear monotonic \cite{snelson2004warped}.
In addition, several previous works integrate this idea in modelling as well, and representative examples are deep GPs \cite{DBLP:journals/corr/DaiDGL15} and hierarchical GPs \cite{DBLP:journals/corr/TranRB15}. 

\section{Neural Process under Doubly Stochastic Variational Inference}
In the last section, we gain more insights about mechanism of GPs and NPs and disentangle these models with the implicit LVM. A conclusion can be drawn that the posterior inference conditioned on the context requires both \textit{approximate distributions with permutation invariance} and \textit{some bridge to connect observations and the target in latent space}. 
Note that the global induced latent variable may be insufficient to describe dependencies, and critical challenge comes from \textit{non-stationarity} and \textit{locality}, which are even crucial in high-dimensional cases.

Hence, we present a hierarchical way to modify NPs, and the trick is to involve auxiliary latent variables for NPs and derive a new evidence lower bound for different levels of random variables with doubly stochastic variational inference \cite{salimbeni2017doubly,titsias2014doubly}. The original intention of involving auxiliary latent variables is to improve the expressiveness of  approximate posteriors, and it is commonly used in deep generative models \cite{maaloe2016auxiliary}. So, as displayed in Table (\ref{np_table}), DSVNP considers a global latent variable and a local latent variable to convey context information at different levels. Our work is also consistent with the hierarchical implicit Bayesian neural networks \cite{DBLP:journals/corr/TranRB15,tran2017hierarchical}, which distinguish the role of latent variables and induce more powerful posteriors.
Without exception, the local latent variable $z_*$ refers to any data point $(x_*,y_* )$ for prediction in DSVNP in the remainders of this paper.

\subsection{Neural Process with Hierarchical Latent Variables}
To extract hierarchical context information for the predictive distribution, we distinguish the global latent variable and the local latent variable in the form of a Bayesian model, and the induced LVM is DSVNP. This variant shares the same prediction structure with AttnNP. The global latent variable is shared across all observations, and the role of context points resembles inducing points in sparse GP \cite{snelson2006sparse}, summarizing general statistics in the task. As for the local latent variable in our proposed DSVNP, it is an auxiliary latent variable responsible mainly for variations of instance locality. From another perspective, DSVNP combines the global latent variable in vanilla NPs with the local latent variable in conditional variational autoencoder (C-VAE). This implementation in model construction separates the global variations and sample specific variations and theoretically increases the expressiveness of the neural network.

As illustrated in Figure \ref{fig1} (d), the target to predict is governed by these two latent variables. Both the global latent variable $z_G$ and the local latent variable $z_*$ contribute to prediction. Formally, the generative model as a $\mathcal{SP}$ is described as follows, where exact inferences for latent variables $z_G$ and $z_*$ are infeasible.
\begin{equation}\label{eq_10}
\begin{split}
    \rho_{x_{1:N+M}}(y_{1:N+M})=\iint\prod_{i=1}^{N+M}p(y_i\vert z_G,z_i,x_i)\\
    p(z_i\vert x_i,z_G)p(z_G)dz_{1:N+M}dz_G   
\end{split}
\end{equation}
Meanwhile, we emphasize that this generation method naturally induces an \textit{exchangeable stochastic process} \cite{bhattacharya2009stochastic}. (The proof is given in Appendix C.)

\subsection{Approximate Inference and ELBO}
With the relationship between these variables clarified, we can characterize the inference process for DSVNP and then a new ELBO is presented. Distinguished from AttnNP, we need to infer both global and local latent variables with evidence of collected dataset. Posteriors of the global and local latent variables on training dataset are approximated with distributions like vanilla NPs, mapping Eq. (\ref{eq_8}) to means and variances. And inference towards local latent variables requires target information in the approximate posterior,
\begin{equation}\label{eq_11}
    q_{\phi_{1,1}}=\mathcal{N}\big(z_G\vert\mu(x_C,y_C,x_T,y_T),\Sigma(x_C,y_C,x_T,y_T)\big)
\end{equation}
\begin{equation}\label{eq_12}
    q_{\phi_{2,1}}=\mathcal{N}\big(z_*\vert\mu(z_G,x_*,y_*),\Sigma(z_G,x_*,y_*)\big)
\end{equation}
where $q_{\phi_{1,1}}$ and $q_{\phi_{2,1}}$ are approximate posteriors in training process.
The generative process is reflected in Eq. (\ref{eq_13}) for DSVNP, where $g_\theta$ indicates a decoder in a neural network.
\begin{equation}\label{eq_13}
    p(y_*\vert x_C,y_C,x_*)=g_{\theta}(z_G,z_*,x_*)
\end{equation}
Consequently, this difference between vanilla NP and DSVNP leads to another ELBO or negative variational free energy $\mathcal{L}$ as the right term,
\begin{equation}\label{eq_14}
    \begin{split}
        \ln\big[p(y_*\vert x_C,y_C,x_*)\big]\geq \mathbb{E}_{q_{\phi_{1,1}}}\mathbb{E}_{q_{\phi_{2,1}}}\ln[p(y_*\vert z_G,z_*,x_*)] \\
        -\mathbb{E}_{q_{\phi_{1,1}}}[D_{KL}[q_{\phi_{2,1}}(z_*\vert z_G,x_*,y_*)\parallel p_{\phi_{2,2}}(z_*\vert z_G,x_*)]\big] \\
        -D_{KL}\big[q_{\phi_{1,1}}(z_G\vert x_C,y_C,x_T,y_T)\parallel p_{\phi_{1,2}}(z_G\vert x_C,y_C)\big]
    \end{split}
\end{equation}
where $p_{\phi_{1,2}}(z_G\vert x_C,y_C)$ and $p_{\phi_{2,2}}(z_*\vert z_G,x_*)$ parameterized with neural networks are used as prior distributions. Here we no longer employ standard normal distributions with zero prior information, and instead these are parameterized with two diagonal Gaussians for the sake of simplicity and learned in an amortized way.

\subsection{Scalable Training and Uncertainty-aware Prediction}
Based on the inference process in DSVNP and the corresponding ELBO in Eq. (\ref{eq_14}), the Monte Carlo estimation for the lower bound is derived, in which we wish to maximize,
\begin{equation}\label{eq_15}
    \begin{split}
        \mathcal{L}_{MC}=\frac{1}{K}\sum_{k=1}^K\big[\frac{1}{S}\sum_{s=1}^S\ln[p(y_*\vert x_*,z_*^{(s)},z_G^{(k)})] \\
        -D_{KL}[q(z_*\vert z_G^{(k)},x_*,y_*)\parallel p(z_*\vert z_G^{(k)},x_*)]\big] \\
        -D_{KL}\big[q(z_G\vert x_C,y_C,x_T,y_T)\parallel p(z_G\vert x_C,y_C)\big]
    \end{split}
\end{equation}
where latent variables are sampled as $z_G^{(k)}\sim q_{\phi_{1,1}}(z_G\vert x_C,y_C)$ and $z_*^{(s)}\sim q_{\phi_{2,1}}(z_*\vert z_G^{(k)},x_*,y_*)$. And the resulted Eq. (\ref{eq_15}) is employed as the objective function in the training process. To reduce variance in sampling, the reparameterization trick \cite{kingma2014stochastic} is used for all approximate distributions and the model is optimized using Stochastic Gradient Variational Bayes \cite{kingma2014stochastic}. More details can be found in Algorithm (\ref{alg_tr}).

\begin{algorithm}[tb]
   \caption{Variational Inference for DSVNP in Training.}
   \label{alg_tr}
\begin{algorithmic}
   \STATE {\bfseries Input:} Dataset $\mathcal{D}$, Maximum context points $N_{max}$,
   Learning rate $\alpha$, Batch size $B$ and Epoch number $m$.
   \STATE {\bfseries Output:} Model parameters $\phi_1$, $\phi_2$ and $\theta$.
   \STATE Initialize parameters $\phi_1$, $\phi_2$, $\theta$ of recognition model and generative model in Eq. (\ref{eq_11}), (\ref{eq_12}) and (\ref{eq_13}).
   \FOR{$i=1$ {\bfseries to} $m$}
   \STATE Draw some context number $N_C\sim U[1,N_{max}]$;
   \STATE Draw mini-batch instances and formulate context-target pairs $\{(x_C,y_C,x_T,y_T)_{bs}\}_{bs=1}^B\sim \mathcal{D}$;
   \STATE Feedforward the mini-batch instances to recognition model $q_{\phi_1}$:
   \STATE\qquad Draw sample of latent variable $z_G\sim q_{\phi_{1,1}}$ in 
   \STATE\qquad Eq. (\ref{eq_11});
   \STATE\qquad Draw sample of latent variable $z_*\sim q_{\phi_{2,1}}$ in 
   \STATE\qquad Eq. (\ref{eq_12});
   \STATE Feedforward the latent variables to discrimination model $p_{\theta}$:
   \STATE\qquad Compute conditional probability distribution in 
   \STATE\qquad Eq. (\ref{eq_13});
   \STATE Update parameters by Optimizing Eq. (\ref{eq_15}):
   \STATE\qquad $\phi_1\gets\phi_1+\alpha\nabla_{\phi_1}\mathcal{L}_{MC}$\; $\triangleright$\;$\phi_1=[\phi_{1,1},\phi_{1,2}]$
   \STATE\qquad $\phi_2\gets\phi_2+\alpha\nabla_{\phi_2}\mathcal{L}_{MC}$\; $\triangleright$\;$\phi_1=[\phi_{2,1},\phi_{2,2}]$
   \STATE\qquad $\theta\gets\theta+\alpha\nabla_{\theta}\mathcal{L}_{MC}$
   \ENDFOR
\end{algorithmic}
\end{algorithm}

The predictive distribution is of our interest. For DSVNP, prior networks as $p(z_G\vert x_C,y_C)$ and $p(z_*\vert z_G,x_*)$ are involved in prediction, and this leads to the integration over both global and local latent variables here as revealed in Eq. (\ref{eq_16}).
\begin{equation}\label{eq_16}
    \begin{split}
        p(y_*\vert x_C,y_C,x_*)=\iint
        p(y_*\vert z_G,z_*,x_*)p_{\phi_{1,2}}(z_G\vert x_C,y_C) \\
        p_{\phi_{2,2}}(z_*\vert z_G,x_*)dz_{G}dz_*
    \end{split}
\end{equation}
For uncertainty-aware prediction, there exist different approaches for Bayesian neural networks. Generally, once the model is well trained, the conditional distribution in neural networks can be derived. The accuracy can be evaluated through deterministic inference over latent variables, i.e., $\tilde{z_G}=\mathbb{E}[z_G\vert x_C,y_C]$, $\tilde{z_*}=\mathbb{E}[z_*\vert z_G,x_*]$, $y_*=\arg \max_{y} p(y\vert \tilde{z_*},\tilde{z_G,x_*})$.

The Monte Carlo estimation over Eq. (\ref{eq_16}), which is commonly used for prediction, can be written in the following equation,
\begin{equation}\label{eq_17}
    p(y_*\vert x_C,y_C,x_*)\approx \frac{1}{KS}\sum_{k=1}^K\sum_{s=1}^S p_{\theta}(y_*\vert x_*,z_*^{(s)},z_G^{(k)})
\end{equation}
where the global and local latent variables are sampled in prior networks through ancestral sampling as $z_G^{(k)}\sim p_{\phi_{1,2}}(z_G\vert x_C,y_C)$ and $z_*^{(s)}\sim p_{\phi_{2,2}}(z_*\vert z_G^{(k)},x_*)$.

\subsection{More Insights and Implementation Tricks}
The global latent variable and local latent variables govern different variations in prediction and sample generation. This is a part of motivations for AttnNP and DSVNP. Interestingly, the inference for our induced $\mathcal{SP}$ integrates the aspects of vanilla NPs \cite{eslami2018neural} and C-VAEs \cite{sohn2015learning}.

Similar to $\beta$-VAE \cite{higgins2017beta}, we rewrite the right term in Eq. (\ref{eq_14}) with constraints and these restrict the search for variational distributions. Equivalently, tuning the weights of divergence terms in Eq. (\ref{eq_14}) leads to varying balance between global and local information.
\begin{equation}\label{eq_18}
    \begin{split}
        \max_{\phi_1,\phi_2,\theta} \mathbb{E}_{q_{\phi_{1,1}}}\mathbb{E}_{q_{\phi_{2,1}}}\ln[p_{\theta}(y_*\vert z_G,z_*,x_*)] \\
        D_{KL}\big[q(z_G\vert x_C,y_C,x_T,y_T)\parallel p(z_G\vert x_C,y_C)\big]<\epsilon_{G} \\
        \mathbb{E}_{q_{\phi_{1,1}}}\big[D_{KL}[q(z_*\vert z_G,x_*,y_*)\parallel p(z_*\vert z_G,x_*)]\big]<\epsilon_{L}
    \end{split}
\end{equation}
Here, a more practical objective in implementations derived by weight calibrations in Eq. (\ref{eq_19}).
\begin{equation}\label{eq_19}
    \begin{split}
        \mathcal{L}_{MC}^{W}=\frac{1}{K}\sum_{k=1}^K\big[\frac{1}{S}\sum_{s=1}^S\ln[p(y_*\vert x_*,z_*^{(s)},z_G^{(k)})] \\
        -\beta_{1}D_{KL}[q(z_*\vert z_G^{(k)},x_*,y_*)\parallel p(z_*\vert z_G^{(k)},x_*)]\big] \\
        -\beta_{2}D_{KL}\big[q(z_G\vert x_C,y_C,x_T,y_T)\parallel p(z_G\vert x_C,y_C)\big]
    \end{split}
\end{equation}
Also, training stochastic model with multiple latent variables is non-trivial, and there exist several works about KL divergence term annealing \cite{sonderby2016train} or dynamically adapting for the weights. Importantly, the target specific KL divergence term is sometimes suggested to assign more penalty to guarantee the consistency between approximate posterior and prior distribution \cite{kohl2018probabilistic,sohn2015learning}. 

\begin{figure}[ht]
\vskip 0.1in
\begin{center}
\centerline{\includegraphics[width=0.5\textwidth,height=0.2\textheight]{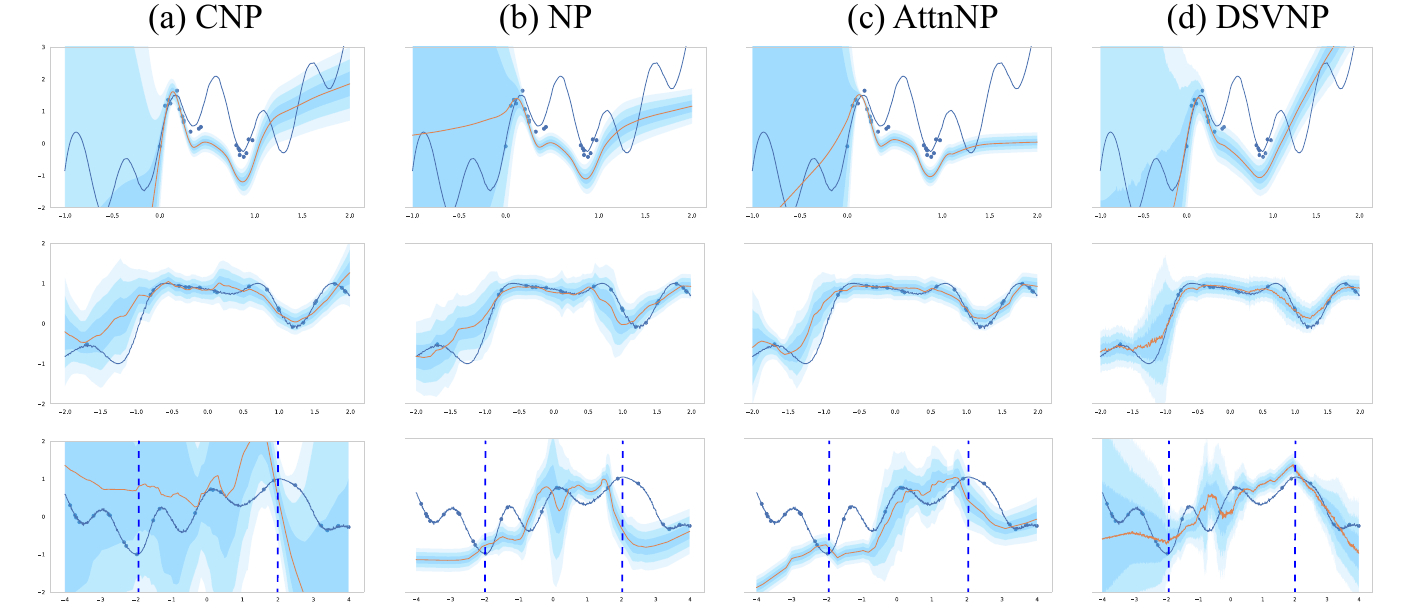}}
\caption{Function Prediction in Interpolation and Extrapolation. Blue curves are ground truth with dotted units as context points, and orange ones are predicted results. Rows from up to down respectively indicate cases: single function with noise, interpolation and extrapolation towards realizations from stochastic process. The shadow regions are $\pm$3 standard deviations from the mean.}
\label{fig2}
\end{center}
\vskip -0.1in
\end{figure}

\section{Experiments}
In this section, we start with learning predictive functions on several toy dataset, and then high-dimensional tasks, including system identification on physics engines, multioutput regression on real-world dataset as well as image classification with uncertainty quantification, are performed to evaluate properties of NP related models. The dot-product attention is implemented in all AttnNPs here. All implementation details are attached in Appendix E.

\subsection{Synthetic Experiments}
We initially investigate the \textit{episdemic} uncertainty captured by NP related models on a 1-d regression task, and the function \cite{osband2016deep} is characterized as $y=x+\epsilon+\sin(4(x+\epsilon))+\sin(13(x+\epsilon))$. Observations as the training set include 12 points and 8 points respectively uniformly drawn from intervals $U[0,0.6]$ and $U[0.8,1.0]$, with the noise drawn from $\epsilon\sim N(0,0.003^2)$. As illustrated in the first row of Fig. (\ref{fig2}), we can observe CNP and DSVNP better quantify variance outside the interval $[0,1.0]$, while AttnNP either overestimates or underestimates the uncertainty to show higher or lower standard deviations in regions with less observations. All models share similar properties with GPs in predictive distributions, displaying lower variances around observed points. As for the gap in interval $[0.6,0.8]$, the revealed uncertainty is consistent to that in \cite{DBLP:conf/iclr/SunZSG19,hernandez2015probabilistic} with intermediate variances. 

Further, we conduct curve fitting tasks in $\mathcal{SP}$. The $\mathcal{SP}$ initializes with a zero mean Gaussian Process $y^{(0)}\sim \mathcal{GP}(0,k(.,.))$ indexed in the interval $x\in[-2.0,2.0]$, where the radial basis kernel $k(x,x^{\prime})=\sigma^{2}exp(-(x-x^{\prime})^{2}/2l^2)$ is used with \textit{l}-1 0.4 and $\sigma$ 1.0. Then the transformation is performed to yield $y=\sin{(y^{(0)}(x)+x)}$. The training process follows that in NP \cite{garnelo2018neural}.
Predicted results are visualized in the second and the third rows of Fig. {(\ref{fig2})}. Note that CNP only predicts points out of the context in default settings. More evidence is reported in Table (\ref{sim_results}), where 2000 realizations are independently sampled and predicted for both interpolation and extrapolation. After several repetitive observations, we find in terms of the interpolation accuracy, DSVNP works better than vanilla NP but the improvement is not as significant as that in AttnNP, which is also verified in visualizations. All (C)NPs show higher uncertainties around index 0, where less context points are located, and variances are relatively close in other regions. For extrapolation results, since all models are trained in the dotted column lines restricted regions, it is tough to scale to regions out of training interval and all negative log-likelihoods (NLLs) are higher. When there exist many context points located outside the interval, the learned context variable may deteriorate predictions for all (C)NPs, and observations confirm findings in \cite{DBLP:conf/iclr/0003BFRDT20}. Interestingly, DSVNP tends to overestimate uncertainties out of the training interval but predicted extrapolation results mostly fall into the one $\sigma$ confident region, this property is similar to CNP. On the other hand, vanilla NP and AttnNP tend to underestimate the uncertainty sometimes. 

\begin{table}[t]
\caption{Average Negative Log-likelihoods over all target points on realizations from Synthetic Stochastic Process. (Figures in brackets are variances.)}
\label{sim_results}
\vskip 0.15in
\begin{center}
\begin{small}
\begin{sc}
\begin{tabular}{llllll}
\toprule
Prediction &CNP &NP &AttnNP &DSVNP \\
\midrule
Inter &-0.802 &-0.958 &\textbf{-1.149} &-0.975 \\
      &(1e-6) &(2e-5) &\textbf{(8e-6)} &(2e-5) \\
Extra &\textbf{1.764} &8.192 &8.091 &\textbf{4.203} \\
      &\textbf{(1e-1)} &(7e1) &(7e2) &\textbf{(9e0)} \\
\bottomrule
\end{tabular}
\end{sc}
\end{small}
\end{center}
\vskip -0.2in
\end{table}

\subsection{System Identification on Physics Engines}
\begin{figure}[ht]
\vskip 0.1in
\begin{center}
\centerline{\includegraphics[width=0.25\textwidth,height=0.15\textheight]{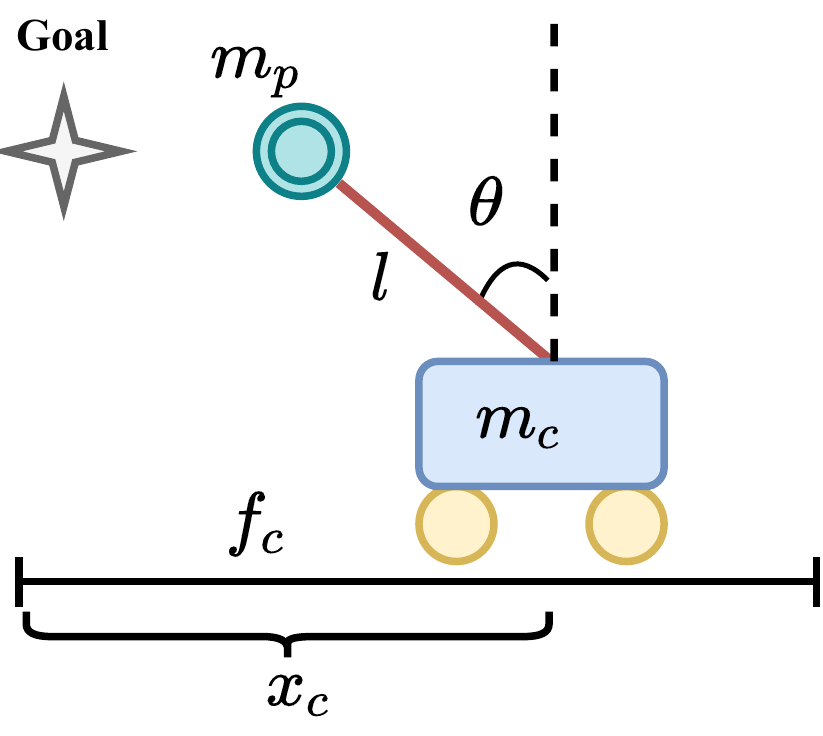}}
\caption{Cart-Pole Dynamical Systems.The cart and the pole are with masses $m_c$ and $m_p$, and the length of the pole is $l$. And the configuration of the simulator is up to parameters of the cart-pole mass and the ground friction coefficient here with other hyper-parameters fixed in this experiment.}
\label{fig4}
\end{center}
\vskip -0.3in
\end{figure}

Capturing dynamics in systems is crucial in control related problems, and we extend synthetic experiments on a classical simulator, Cart-Pole systems, which is detailed in \cite{gal2016improving}. As shown in Fig. (\ref{fig4}), the original intention is to conduct actions to reach the goal with the end of a pole, but here we focus on dynamics and the state is a vector of the location, the angle and their first-order derivatives.
Specifically, the aim is to forecast the transited state $[x_c,\theta,x_c^{\prime},\theta^{\prime}]$ in time step $t+1$ based on the input as a state action pair $[x_c,\theta,x_c^{\prime},\theta^{\prime},a]$ in time step $t$. 
To generate a variety of trajectories under a random policy for this experiment, the mass $m_c$ and the ground friction coefficient $f_c$ are varied in the discrete choices $m_c\in\{0.3,0.4,0.5,0.6,0.7\}$ and $f_c\in\{0.06,0.08,0.1,0.12\}$. Each pair of $[m_c,f_c]$ values specifies a dynamics environment, and we formulate all pairs of $m_c\in\{0.3,0.5,0.7\}$ and $f_c\in\{0.08,0.12\}$ as training environments with the rest 16 pairs of configurations as the testing environments.

For each configuration of the simulator including training and testing environments, we sample 400 trajectories of horizon as 10 steps using a random controller, and more details refer to Appendix E. The training process follows Algorithm (\ref{alg_tr}) with the maximum number of context points as 100. During the testing process, 100 state transition pairs are randomly selected for each configuration of the environment, working as the maximum context points to identify the configuration of dynamics. And the collected results are reported in Table (\ref{cartpole_results}), where prediction performance on 5600 trajectories from 14 configurations of environments are revealed. As can be seen, the negative log-likelihood values are not consistent with those of mean square errors, and DSVNP shows both better uncertainty quantification with lowest NLLs and approximation errors in MSEs. AttnNP improves NP in both metrics, while CNP shows relatively better NLLs but the approximation error is a bit higher than others. 

\begin{table}[t]
\caption{Predictive Negative Log-Likelihoods and Mean Square Errors on Cart-Pole State Transition Testing Dataset. (Figures in brackets are variances.)}
\label{cartpole_results}
\vskip 0.1in
\begin{center}
\begin{small}
\begin{sc}
\begin{tabular}{lllll}
\toprule
Metrics &CNP &NP &AttnNP &DSVNP\\
\midrule
NLL &-2.014  &-1.537 &-1.821 &\textbf{-2.145} \\
&(9e-4)  &(1e-3) &(7e-3) &\textbf{(9e-4)} \\
MSE &0.096 &0.074 &0.067 &\textbf{0.036} \\
&(3e-4) &(2e-4) &(1e-4) &\textbf{(2.1e-5)} \\
\bottomrule
\end{tabular}
\end{sc}
\end{small}
\end{center}
\vskip -0.2in
\end{table}

\subsection{Multi-Output Regression on Real-world Dataset}
Further, more complicated scenarios are considered when the regression task relates to multiple outputs. As investigated in \cite{moreno2018heterogeneous,bonilla2008multi}, distributions of output variables are implicit, which means no explicit distributions are appropriate to be used in parameterizing the output. 
We evaluate the performance of all models on dataset, including SARCOS \footnote{http://www.gaussianprocess.org/gpml/data/}, Water Quality (WQ) \cite{dvzeroski2000predicting} and SCM20d \cite{spyromitros2016multi}. Details about these dataset and neural architectures for all models are included in Appendix E. Furthermore, Monte-Carlo Dropout is included for comparisons. Similar to NP \cite{garnelo2018neural}, the variance parameter is not learned and the objective in optimization is pointwise mean square errors (MSEs) after averaging all dimensions in the output. Each dataset is randomly split into 2-folds as training and testing sets. The training procedure in (C)NPs follows that in Algorithm (\ref{alg_tr}), and some context points are randomly selected in batch samples. In the testing stage, we randomly select 30 instances as the context and then perform predictions with (C)NPs. The weights of data likelihood and KL divergence terms in models are not tuned here.

\begin{table*}[t]
\caption{Predictive MSEs on Multi-Output Dataset. CNP's results are for target points. $D$ records (input,output) dimensions, and $N$ is the number of samples. MC-Dropout runs 50 stochastic forward propagation and average results for prediction in each data point. (Figures in brackets are variances.)}
\label{mulregression_results}
\vskip 0.2in
\begin{center}
\begin{small}
\begin{sc}
\begin{tabular}{llllllll}
\toprule
Dataset &$N$ &$D$ &MC-Dropout &CNP &NP &AttnNP &DSVNP\\
\midrule
Sarcos &48933 &(21,7) &1.215(3e-3) &1.437(2.9e-2) &1.285(1.2e-1) &1.362(8.4e-2) &\textbf{0.839(1.5e-2)} \\
WQ &1060 &(16,14) &0.007(9.6e-8) &0.015(2.4e-5) &0.007(5.2e-6) &0.01(8.5e-6)  &\textbf{0.006(1.6e-6)} \\
SCM20d &8966 &(61,16) &0.017(2.4e-7)  &0.037(6.7e-5)  &0.015(7.1e-8)  &0.015(8.1e-7)  &\textbf{0.007(2.3e-7)} \\
\bottomrule
\end{tabular}
\end{sc}
\end{small}
\end{center}
\vskip -0.2in
\end{table*}

During training, ELBOs in NP related models are optimized, while MSEs are used as evaluation metric in testing \cite{dezfouli2015scalable}\footnote{Directly optimizing Gaussian log-likelihoods does harm to performance based on experimental results.}. The predictive results on testing dataset are reported in Table (\ref{mulregression_results}). All MSEs are averaged after 10 independent experiments. We observe DSVNP outperforms other models, and deterministic context information in CNP hardly increases performance. Compared with NP models, MC-NN is relatively satisfying on Sarcos and WQ, and AttnNP works not well in these cases. A potential reason can be that deterministic context embedding with dot product attention is less predictive for output with multiple dimensions, while the role of local latent variable in DSVNP not only bridges the gap between input and output, but also extracts some correlation information among variables in outputs. As a comparison to synthetic experiments, the attention mechanism is more suitable to extract local information when the output dimension is lower. 

\begin{figure*}[ht]
\vskip 0.2in
\begin{center}
\centerline{\includegraphics[width=1.2\textwidth,height=0.22\textheight]{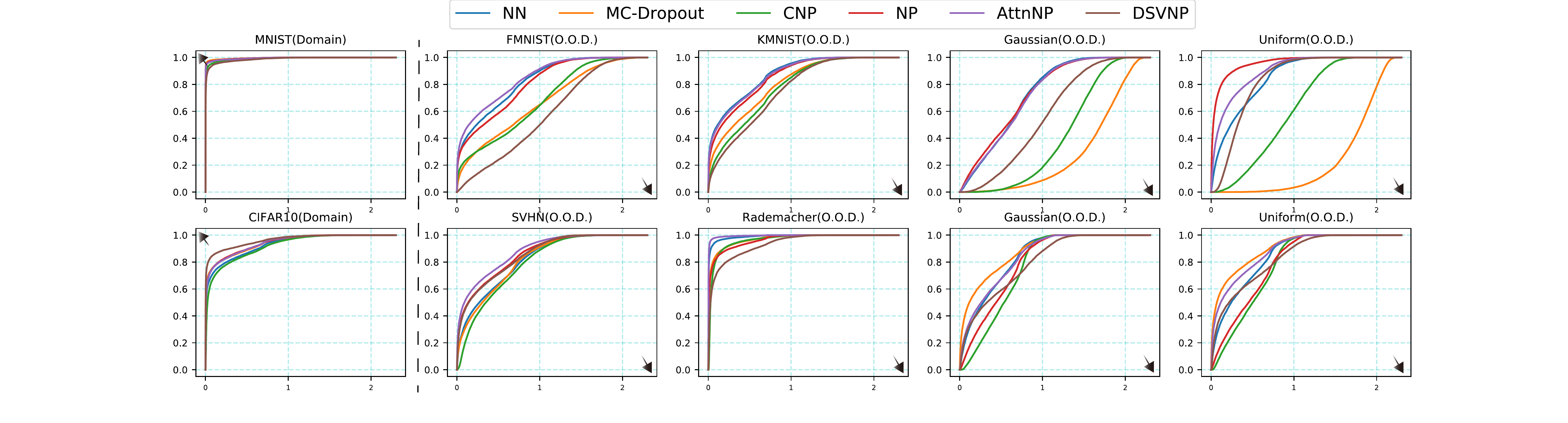}}
\caption{Cumulative Distribution Functions of Entropies in O.O.D. Detection Tasks. Values in X-axis are example entropies ranging from 0 to 2.3, and y-axis records cumulative probabilities. The first row corresponds to the predictive result with models trained on MNIST, while the second is with models trained on CIFAR10. NN means the baseline neural network without dropout layer. Curves in CDFs closer to triangular arrows reveal better uncertainty quantification.}
\label{fig3}
\end{center}
\vskip -0.3in
\end{figure*}

\subsection{Classification with Uncertainty Quantification}
Here image classification is performed with NP models and MC-Dropout, and out of distribution (o.o.d.) detection is chosen to measure the goodness of uncertainty quantification. We respectively train models on MNIST and CIFAR10. The dimensions for latent variables are 64 on MNIST and 128 on CIFAR10. The training process for NP related models follows Algorithm (\ref{alg_tr}), with the number of context images randomly selected in each batch update. For the testing process, we randomly select 100 instances from the domain dataset as the context for (C)NP models. The commonly used measure for uncertainty in K-class o.o.d. detection is entropy \cite{lakshminarayanan2017simple}, $\mathbb{H}[y^{*}\vert x^{*}]=-\Sigma_{c=1}^{K}p_{w}(y^{*}_c\vert x^{*};\mathcal{D}_{tr})\ln p_{w}(y^{*}_c\vert x^{*};\mathcal{D}_{tr})$, where data point $(x^*,y^*)$ comes from either domain test dataset $\mathcal{D}_{te}$ or o.o.d dataset $\mathcal{D}_{ood}$.

For classification performance with NP related models, we observe the difference is extremely tiny on MNIST with all accuracies around 99\%, while on CIFAR10 DSVNP beats all baselines with highest accuracy 86.3\% and lowest in-distribution entropies (Refer to Table (2) in Appendix E). The involvement of a deterministic path does not improve much, and in contrast, MC-Dropout and CNP achieve intermediate performance. A possible cause can be implicit kernel information captured by attention network in images is imprecise. The cumulative distributions about predictive entropies are reported in Fig. (\ref{fig3}). For models trained on MNIST, we observe no significant difference on domain dataset, but DSVNP achieves best results on FMNIST/KMNIST and MC-Dropout performs superior on Uniform/Gaussian noise dataset. Interestingly, AttnNP tends to underestimate uncertainty on FMNIST/KMNIST and the measure is close to the neural network without dropout. Those trained on CIFAR10 are are different from observations in the second row of Fig. (\ref{fig3}). It can be noticed DSVNP shows lowest uncertainty on domain dataset (CIFAR10) and medium uncertainty on SVHN/Gaussian/Uniform Dataset. MC-Dropout and AttnNP seem to not work so well overall, but CNP well measures uncertainty on Gaussian/Uniform dataset. Results again verify SVHN as tough dataset for the task \cite{DBLP:conf/iclr/NalisnickMTGL19}. Also note that entropy distributions on Rademacher Dataset are akin to that on domain dataset, which means the Rademacher noise is more risky for CIFAR10 classification, and DSVNP is a better choice to avoid such adversarial attack in this case. 
Those evidences tell us that the deterministic path in AttnNP does not boost classification performance on domain dataset but weakens the ability of o.o.d. detection mostly, while local latent variables in DSVNP improve both performance. Maybe deterministic local latent variables require more practical attention information, but here only dot-product attention information is included. As a comparison, the local latent variable in DSVNP captures some target specific information during the training process and improves detection performance with it.

\section{Related Works}
\textbf{Scalability and Expressiveness in Stochastic Process.} GPs are the most well known member of $\mathcal{SP}$s family and have inspired a lot of extensions, such as deep kernel learning \cite{wilson2016deep,wilson2016stochastic} and sparse GPs \cite{snelson2006sparse} with better scalability. Especially, the latter incorporated sparse prior in function distribution and utilized a small proportion of observations in predictions. In multi-task cases, several GP variants were proposed \cite{moreno2018heterogeneous,bonilla2008multi,zhao2016variational}. Other works also achieve sparse effect but with variational inference, approximating the posterior in GPs and optimizing ELBO \cite{hensman2015scalable,salimbeni2019deep,titsias2010bayesian}. Another branch is about directly capturing uncertainties with deep neural networks, which is revealed in NP related models. Other extensions include generative query network \cite{eslami2018neural}, sequential NP \cite{singh2019sequential} and convolutional conditional NP \cite{DBLP:conf/iclr/0003BFRDT20}. 
Variational implicit process \cite{ma2019variational} targeted at more general $\mathcal{SP}$s and utilized GPs in latent space as approximation. 
Sun et al. proposed functional variational Bayesian neural networks \cite{DBLP:conf/iclr/SunZSG19}, and variational distribution over functions of measurement set was used to represent $\mathcal{SP}$s. The more recently proposed functional NPs \cite{louizos2019functional} characterized a novel family of exchangeable stochastic processes, placing more flexible distributions over latent variables and constructing directed acyclic graphs with latent affinities of instances in inference and prediction.

\textbf{Uncertainty Quantification and Computational Complexity.}
GPs can well characterize aleatoric uncertainty and epistemic uncertainty through kernel function and Gaussian noise. But $\mathcal{SP}$s with non-Gaussian marginals are crucial in modelling. Apart from GPs, there exist some other techniques such as Dropout \cite{gal2016dropout} or other variants of Bayesian neural networks \cite{louizos2017bayesian} to quantify uncertainty. In \cite{depeweg2018decomposition}, uncertainties were further decomposed in Bayesian neural network. DSVNP can theoretically capture both uncertainties as an approximate prediction model for general $\mathcal{SP}$s and approaches the problem in a Bayesian way. For the computational cost in prediction, the superior sparse GPs with K-rank covariance matrix approximations \cite{burt2019rates} are with the complexity $\mathcal{O}((M+N)K^2)$, while the variants of CNPs or NPs mostly reduce the complexity $\mathcal{O}((N+M)^3)$ in GPs to $\mathcal{O}(M+N)$ in prediction process. And those for AttnNP and DSVNP are $\mathcal{O}((M+N)N)$.

\section{Discussion and Conclusion}
In this paper, we present a novel exchangeable stochastic process as DSVNP, which is formalized as a latent variable model. DSVNP integrates latent variables hierarchically and improves the expressiveness of the vanilla NP model.  Experiments on high-dimensional tasks demonstrate better capability in prediction and uncertainty quantification. Since this work mainly concentrates on latent variables and associated inference methods, future directions can be the enhancement in the representation of latent variables, such as the use of more flexible equivariant transformations over the context or the dedicated selection of proper context points.

\section*{Acknowledgements}
We express thanks to Christos Louizos and Bastiaan Veeling for helpful discussions. We would like to thank Daniel Worrall, Patrick Forré, Xiahan Shi, Sindy Löwe and anonymous reviewers for precious feedback on the initial manuscript. Q. Wang gratefully acknowledges consistent and sincere support from Prof. Max Welling and was also supported by China Scholarship Council during his pursuing Ph.D at AMLAB. Also, we gratefully acknowledge the support of NVIDIA Corporation with the donation of a Titan V GPU.

\bibliography{example_paper}
\bibliographystyle{icml2020}

\appendix
\onecolumn
\icmltitle{Supplementary Materials}
\renewcommand\thesection{\Alph{section}}
\setcounter{table}{0}

\section{Some Basic Concepts}
At first, we revisit some basics, which would help us understand properties of GPs and NPs. Fundamental concepts include \emph{permutation invariant function} (\emph{PIF}), \emph{general stochastic function process} (\emph{GSFP}) etc. We stress the importance of relationship disentanglement between GPs and NPs and motivations in approximating stochastic processes (\emph{SPs}) in the main passage.

\textbf{Definition 1. Permutation Invariant Function.}
The $\textit{f}$ mapping a set of $N$ $M$-dimension elements to a $D$-dimension vector variable is said to be a \textit{permutation invariant function} if:
\begin{equation}\label{eq_pif}
\begin{split}
f:\mathcal{\times}_{i=1}^{N} \mathcal{R}_{i}^{M}\rightarrow \mathcal{R}^{D} \\
\textit{x}=[x_1,\dots,x_N]\mapsto \textit{f}=\big[f_{1}(x_{\pi(1:N)}),\dots,f_{D}(x_{\pi(1:N)})\big]
\end{split}
\tag{A.1}
\end{equation}
where $x_i$ is a $M$-dimensional vector, $x_{1:N}=[x_1,x_2,\dots,x_N]$ is a set, operation $\pi:[1,2,\dots,N]\mapsto[\pi_1,\pi_2,\dots,\pi_N]$ imposes a permutation over the order of elements in the set. The \textit{PIF} suggests the image of the map is regardless of the element order. Another related concept \textit{permutation equivariant function (PEF)} keeps the order of elements in the output consistent with that in the input under any permutation operation $\pi$.
\begin{equation}\label{eq_2}
\begin{split}
\textit{f}:\mathcal{\times}_{i=1}^{N} \mathcal{R}_{i}^{M}\rightarrow \mathcal{R}^{N} \\
\textit{x}_{\pi}=(x_{\pi_1},\dots,x_{\pi_N})\mapsto\textit{f}_{\pi}=\pi\circ\textit{f}(\textit{x}_{1:N})
\end{split}
\tag{A.2}
\end{equation}

The \textbf{Definition 1} is important, since the exchangeable stochastic function process of our interest in this domain is intrinsically a distribution over set function, and PIF plays as an inductive bias in preserving invariant statistics. 

Permutation invariant functions are candidate functions for learning embeddings of a set or other order uncorrelated data structure $X_{1:N}$, and several examples can be listed in the following forms.
\begin{itemize}
    \item Some structure with mean/summation over the output:
    \begin{equation} \label{eq_1}
    F(X_{\pi(1:N)})=\big(\frac{1}{N}\Sigma_{i=1}^{N} \phi_{1}(x_i),\frac{1}{N}\Sigma_{i=1}^{N} \phi_{2}(x_i),\dots,\frac{1}{N}\Sigma_{i=1}^{N} \phi_{M}(x_i)\big)
    \tag{A.3}
    \end{equation}
    
    \item Some structure with Maximum/Minimum/Top k-th operator over the output (Take maximum operator for example), such as:
    \begin{equation} \label{eq_2}
    F(X_{\pi(1:N)})=\big(max_{i\in\{1,2,\dots,N\}}  \phi_{1}(x_i),max_{i\in\{1,2,\dots,N\}} \phi_{2}(x_i),\dots,max_{i\in\{1,2,\dots,N\}} \phi_{M}(x_i)\big)
    \tag{A.4}
    \end{equation}
    
    \item Some structure with symmetric higher order polynomials or other functions with a symmetry bi-variate function $\phi$:
    \begin{equation} \label{eq_3}
    F(X_{\pi(1:N)})=\big(\Sigma_{i,j=\{1,2,\dots,N\}}  \phi_{1}(x_i,x_j),\Sigma_{i,j=\{1,2,\dots,N\}}  \phi_{2}(x_i,x_j),\dots,\Sigma_{i,j=\{1,2,\dots,N\}} \phi_{M}(x_i,x_j)\big)
    \tag{A.5}
    \end{equation}
\end{itemize}
The invariant property is easy to be verified in these cases, and note that in all settings of NP family in this paper, \textbf{Eq. (\ref{eq_1})} is used only. For the bi-variate symmetric function in \textbf{Eq. (\ref{eq_3})} or other more complicated operators would result in more flexible functional translations, but additional computation is required as well. Some of the above mentioned transformations are instantiations in DeepSet. Further investigations in this domain can be the exploitation of these higher order permutation invariant neural network into NPs since more correlations or higher order statistics in the set can be mined for prediction. Additionally, Set Transformer is believed to be powerful in a permutation invariant representation.

\textbf{Definition 2. General Stochastic Function Process.}
Let $\mathcal{X}$ denote the Cartesian product of some intervals as the index set and let dimension of observations $d\in \mathbb{N}$. For each $k\in \mathbb{N}$ and any finite sequence of distinct indices $x_{1},x_{2},..,x_{k}\in \mathcal{X}$, let $\nu(x_{1},x_{2},..,x_{k})$ be some probability measure over $(\mathcal{R}^{d})^{k}$. Suppose the used measure satisfies Kolmogorov Extension Theorem, then there exists a probability space $(\Omega,\mathcal{F},\mathcal{P})$, which induces a \textit{general stochastic function process (GSFP)} $\mathcal{F}:\mathcal{X}\times\Omega\to \mathcal{R}^d$, keeping the property
$\nu_{(x_{1},x_{2},..,x_{k})}(\mathcal{C}_{1}\times \mathcal{C}_{2}\times ...\times \mathcal{C}_{k})=\mathcal{P}(\mathcal{F}(x_{1})\in C_1,\mathcal{F}(x_{2})\in C_{2},...,\mathcal{F}(x_{k})\in \mathcal{C}_{k})$ for all $x_{i}\in\mathcal{X}$, $d\in \mathbb{N}$ and measurable sets $\mathcal{C}_{i}\in \mathcal{R}^{d}$.

The \textbf{Definition 2} presents an important concept for stochastic processes in high-dimensional cases, and this is a general description for the task to learn in mentioned related works.
This includes but not limited to GPs and characterizes the distribution over the stochastic function family.

\section{Proof of Proposition 1}
As we know, a Gaussian distribution is closed under marginalization, conditional probability computation and some other trivial operations. Here the statistical parameter invariance towards the order of the context variables in per sample predictive distribution would be demonstrated.

Given a multivariate Gaussian as the context $X=[x_1,x_2,\dots,x_N]^T\sim \mathcal{N}(X;[\mu_1,\mu_2,\dots,\mu_N]^T,
\Sigma(x_1,x_2,\dots,x_N))$, and for any permutation operation over the order $\pi:[1,2,\dots,N]\to [\pi_1,\pi_2,\dots,\pi_N]$, there exist a permutation matrix $\mathcal{P_{\pi}}=[e_{\pi_1}^T,e_{\pi_2}^T,\dots,e_{\pi_N}^T]$, where only the $\pi_i$-th position is one with the rest zeros. Naturally, it results in a permutation over the random variables in coordinates.
\begin{equation} \label{eq_4}
    P_{\pi}[x_1,x_2,\dots,x_N]^T=[x_{\pi_1},x_{\pi_2},\dots,x_{\pi_N}]^T=X_{\pi}
    \tag{B.1}
\end{equation}
The random variable $X_{\pi}$ follows another multivariate Gaussian as $X_{\pi}\sim \mathcal{N}(X_{\pi};P_{\pi}\mu,P_{\pi}\Sigma P_{\pi}^T)=\mathcal{N}(X_{\pi};\mu_{\pi},\Sigma_{\pi})$. In an elementwise way, we can rewrite the statistics in the form as follows.
\begin{gather*} \label{eq_5}
\begin{split}
      &\mathbb{E}[x_{\pi_i}]=\mu_{\pi_i} \\ &\sigma_{ls}^{\pi}=e_{l}^{T}\Sigma_{\pi}e_{s}=e_{l}^{T}P_{\pi}\Sigma P_{\pi}^T e_{s}=e_{\pi_l}\Sigma e_{\pi_s}=cov(x_{\pi_l},x_{\pi_s})  
\end{split}
    \tag{B.2}
\end{gather*}
Notice in \textbf{Eq. (\ref{eq_5})}, the statistics are \textit{permutation equivariant} now.

As the most important component in GPs, the predictive distribution conditioned on the context $D=[X_{1:N},Y_{1:N}]$ can be analytically computed once GPs are well trained and result in some mean function $m_{\theta}$.
\begin{equation} \label{eq_6}
\begin{split}
    &p(y_{*}\vert Y_{1:N})=\mathcal{N}(y_{*}\vert \tilde{\mu},\tilde{\sigma}^2) \\
    &\tilde{\mu}=m_{\theta}(x_{*})+\Sigma_{x_{*},D}\Sigma_{D,D}^{-1} (y_D-m_{\theta}(x_D)) \\
    &\tilde{\sigma}^2=\sigma_{x_*,x_*}^2-\Sigma_{x_*,D}\Sigma_{D,D}\Sigma_{D,x_*}  
\end{split}
    \tag{B.3}
\end{equation}
Similarly, after imposing a permutation $\pi$ over the order of elements in the context, we can compute the first and second order of statistics between $D_{\pi}=[X_{\pi(1:N)},Y_{\pi(1:N)}]$ and per target point $[x_*,y_*]$.
\begin{equation} \label{eq_7}
\begin{split}
    &\Sigma_{x_*,D_{\pi}}=\Sigma_{x_*,D} P_{\pi}^T=P_{\pi}\Sigma_{D,x_*} \\
    &\Sigma_{D_{\pi},D_{\pi}}^{-1}=P_{\pi}\Sigma_{D,D}^{-1}P_{\pi}^T \\
    &y_{D_{\pi}}-m_{\theta}(x_{D_{\pi}})=P_{\pi}(y_D-m_{\theta}(x_D))  
\end{split}
    \tag{B.4}
\end{equation}
Hence, with the property of orthogonality of permutation matrix $P_{\pi}$, it is easy to verify the \textit{permutation invariance} in statistics for per target predictive distribution.
\begin{equation} \label{eq_8}
\begin{split}
    &\Sigma_{x_{*},D}\Sigma_{D,D}^{-1} (y_D-m_{\theta}(x_D))=\Sigma_{x_{*},D_{\pi}}\Sigma_{D_{\pi},D_{\pi}}^{-1} (y_{D_{\pi}}-m_{\theta}(x_{D_{\pi}})) \\
    &\Sigma_{x_*,D}\Sigma_{D,D}\Sigma_{D,x_*}=\Sigma_{x_*,D_{\pi}}\Sigma_{D_{\pi},D_{\pi}}\Sigma_{D_{\pi},x_*}  
\end{split}
    \tag{B.5}
\end{equation}
To inherit such a property, NP employs a permutation invariant function in embeddings, and the predictive distribution in NP models is invariant to the order of context points. Also, when there exist multiple target samples in the predictive distribution, it is trivial that the statistics between the context and the target in a GP predictive distribution are \textit{permutation equivariant} in terms of the order of target variables.

\section{Proof of DSVNP as Exchangeable Stochastic Process}
In the main passage, we formulate the generation of DSVNP as:
\begin{equation}\label{eq_esp}
\begin{split}
    \rho_{x_{1:N+M}}(y_{1:N+M})=\iint\prod_{i=1}^{N+M}p(y_i\vert z_G,z_i,x_i)p(z_i\vert x_i,z_G)p(z_G)dz_{1:N+M}dz_G
\end{split}
\tag{C.1}
\end{equation}
which indicates the scenario of any finite collection of random variables in \textbf{\textit{y}}-space. Our intention is to show this induces an exchangeable stochastic process. Equivalently, two conditions for Kolmogorov Extension Theorem are required to be satisfied.
\begin{itemize}
    \item \textbf{Marginalization Consistency.} Generally, when the integral is finite, the swap of orders in integration is allowed. Without exception, \textbf{Eq.} \textbf{(\ref{eq_esp})} is assumed to be bounded with some appropriate distributions. Then, for the subset of indexes $\{N+1,N+2,\dots,N+M\}$ in random variables \textbf{\textit{y}}, we have:
    \begin{equation}\label{eq_mc}
    \begin{split}
    \int\rho_{x_{1:N+M}}(y_{1:N+M})dy_{N+1:N+M}=\iiint\prod_{i=1}^{N+M}p(y_i\vert z_G,z_i,x_i)\\
    p(z_i\vert x_i,z_G)p(z_G)dz_{1:N+M}dz_Gdy_{N+1:N+M}\\
    =\iint\prod_{i=1}^{N}p(y_i\vert z_G,z_i,x_i)p(z_i\vert x_i,z_G)\big[\iint\prod_{i=N+1}^{N+M}p(y_i\vert z_G,z_i,x_i)p(z_i\vert x_i,z_G)\\
    dy_{N+1:N+M}dz_{N+1:N+M}\big]p(z_G)dz_Gdz_{1:N}\\
    =\iint\prod_{i=1}^{N}p(y_i\vert z_G,z_i,x_i)p(z_i\vert x_i,z_G)p(z_G)dz_{1:N}dz_G=\rho_{x_{1:N}}(y_{1:N})
    \end{split}
    \tag{C.2}
    \end{equation}
    hence, the marginalization consistency is verified.
    \item \textbf{Exchangeability Consistency.} For any permutation $\pi$ towards the index set $\{1,2,\dots,N\}$, we have:
    \begin{equation}\label{eq_ec}
    \begin{split}
    \rho_{x_{1:N}}(y_{1:N})=\iint\prod_{i=1}^{N}p(y_i\vert z_G,z_i,x_i)p(z_i\vert x_i,z_G)p(z_G)dz_{1:N}dz_G\\
    =\iint\prod_{i=1}^{N}\big[p(y_{\pi_i}\vert z_G,z_{\pi_i},x_{\pi_i})p(z_{\pi_i}\vert x_{\pi_i},z_G)dz_{\pi_i}\big]p(z_G)dz_G\\
    =\iint\prod_{i=1}^{N}p(y_{\pi_i}\vert z_G,z_{\pi_i},x_{\pi_i})p(z_{\pi_i}\vert x_{\pi_i},z_G)p(z_G)dz_{\pi_{(1:N)}}dz_G=\rho_{x_{\pi(1:N)}}(y_{\pi(1:N)})
    \end{split}
    \tag{C.3}
    \end{equation}
    hence, the exchangeability consistency is demonstrated as well.
\end{itemize}
With properties in \textbf{Eq.} \textbf{(\ref{eq_mc})} and \textbf{(\ref{eq_ec})}, our proposed DSVNP is an exchangeable stochastic process in this case.

\section{Derivation of Evidence Lower Bound with Doubly Stochastic Variational Inference}
Akin to vanilla NPs, we assume the existence of a global latent variable $z_G$, which captures summary statistics consistent between the context $[x_C,y_C]$ and the complete target $[x_T,y_T]$. With the involvement of an approximate distribution $q\big(z_G\vert [x_C,y_C,x_T,y_T]\big)$, we can naturally have an initial ELBO in the following form.
\begin{equation} \label{eq_9}
\begin{split}
    &\ln\big[p(y_*\vert x_C,y_C,x_*)\big]=\ln\big[\mathbb{E}_{q(z_G\vert x_C,y_C,x_T,y_T)} p(y_*\vert z_G,x_*)\frac{p(z_G\vert x_C,y_C)}{q(z_G\vert x_C,y_C,x_T,y_T)}\big] \\
    &\geq \mathbb{E}_{q(z_G\vert x_C,y_C,x_T,y_T)}\ln\big[p(y_*\vert z_G,x_*)\big]-D_{KL}\big[q(z_G\vert x_C,y_C,x_T,y_T)\parallel p(z_G\vert x_C,y_C)\big]
\end{split}
    \tag{D.1}
\end{equation}
Note that in \textbf{Eq. (\ref{eq_9})}, the conditional prior distribution $p(z_G\vert x_C,y_C)$ is intractable in practice and the approximation is used here and such a prior is employed to infer the global latent variable in testing processes. For the approximate posterior $q(z_G\vert x_C,y_C,x_T,y_T)$, it makes use of the context and the full target information, and the sample $[x_*,y_*]$ is just an instance in the full target.

Further, by introducing a target specific local latent variable $z_*$, we can derive another ELBO for the prediction term in the right side of \textbf{Eq. (\ref{eq_9})} with the same trick.
\begin{gather*}
\begin{split} \label{eq_10}
    \mathbb{E}_{q(z_G\vert x_C,y_C,x_T,y_T)}\ln\big[p(y_*\vert z_G,x_*)\big] \\
   =\mathbb{E}_{q(z_G\vert x_C,y_C,z_T,y_T)}\ln\big[\mathbb{E}_{q(z_*\vert z_G,[x_*,y_*])} p(y_*\vert &z_G,z_*,x_*)\frac{p(z_*\vert z_G,x_*)}{q(z_*\vert z_G,[x_*,y_*])}\big]\geq \\
    \mathbb{E}_{q(z_G\vert x_C,y_C,x_T,y_T)}\mathbb{E}_{q(z_*\vert z_G,[x_*,y_*])}\ln\big[p(y_*\vert z_G,z_*,x_*)\big] \\
    -\mathbb{E}_{q(z_G\vert x_C,y_C,x_T,y_T)}\big[D_{KL}[q(z_*\vert z_G,[x_*,y_*])\parallel p(z_*\vert z_G,x_*)]\big]
\end{split}
    \tag{D.2}
\end{gather*}

With the combination of \textbf{Eq. (\ref{eq_9})} and \textbf{(\ref{eq_10})}, the final ELBO $\mathcal{L}$ as the right term in the following is formulated.
\begin{equation}\label{eq_11}
    \begin{split}
        &\ln\big[\underbrace{p(y_*\vert x_C,y_C,x_*)}_{implicit\;data\;likelihood}\big]\geq \mathbb{E}_{q_{\phi_{1}(z_G)}}\mathbb{E}_{q_{\phi_{2}(z_*)}}\ln[\underbrace{p(y_*\vert z_G,z_*,x_*)}_{data\;likelihood}]\\
        &-\mathbb{E}_{q_{\phi_{1}(z_G)}}[D_{KL}[q(z_*\vert z_G,x_*,y_*)\parallel\underbrace{p(z_*\vert z_G,x_*)}_{local\;prior}]\big]-D_{KL}\big[q(z_G\vert x_C,y_C,x_T,y_T)\parallel\underbrace{p(z_G\vert x_C,y_C)}_{global\;prior}\big]
    \end{split}
    \tag{D.3}
\end{equation}
The real data likelihood is generally implicit, and the ELBO is an approximate objective. Note that the conditional prior distribution in \textbf{Eq. (\ref{eq_11})}, $p(z_*\vert z_G,x_*)$ functions as a local latent variable and is approximated with a Gaussian distribution for the sake of easy implementation. With reparameterization trick, used as: $z_G=\mu_{\phi_1}+\epsilon_{1}\sigma_{\phi_{1}}$ and $z_*=\mu_{\phi_2}+\epsilon_{2}\sigma_{\phi_{2}}$, we can estimate the gradient towards the sample $(x_*,y_*)$ analytically in \textbf{Eq. (\ref{eq_12})}, \textbf{(\ref{eq_13})} and \textbf{(\ref{eq_14})}.
\begin{equation}\label{eq_12}
    \begin{split}
        \frac{\partial\mathcal{L}}{\partial\phi_{1}}=\mathbb{E}_{\epsilon_{1}\sim N(0,I)}\mathbb{E}_{\epsilon_{2}\sim N(0,I)}\frac{\partial}{\partial\phi_{1}}\ln\big[p(y_*\vert \mu_{\phi_1}+\epsilon_{1}\sigma_{\phi_{1}},\mu_{\phi_2}+\epsilon_{2}\sigma_{\phi_{2}},x_*)\big] \\
        -\mathbb{E}_{\epsilon_{1}\sim N(0,I)}\frac{\partial}{\partial\phi_{1}}D_{KL}\big[q(\mu_{\phi_2}+\epsilon_{2}\sigma_{\phi_{2}}\vert \mu_{\phi_1}+\epsilon_{1}\sigma_{\phi_{1}},x_*,y_*)\parallel p(\mu_{\phi_2}+\epsilon_{2}\sigma_{\phi_{2}}\vert \mu_{\phi_1}+\epsilon_{1}\sigma_{\phi_{1}},x_*)\big] \\
        -\mathbb{E}_{\epsilon_{1}\sim N(0,I)}\frac{\partial}{\partial\phi_{1}}D_{KL}\big[q(\mu_{\phi_1}+\epsilon_{1}\sigma_{\phi_{1}}\vert x_C,y_C,x_T,y_T)\parallel p(\mu_{\phi_1}+\epsilon_{1}\sigma_{\phi_{1}}\vert x_C,y_C)\big]
    \end{split}
    \tag{D.4}
\end{equation}

\begin{equation}\label{eq_13}
    \begin{split}
        \frac{\partial\mathcal{L}}{\partial\phi_{2}}=\mathbb{E}_{\epsilon_{1}\sim N(0,I)}\mathbb{E}_{\epsilon_{2}\sim N(0,I)}\frac{\partial}{\partial\phi_{2}}\ln\big[p(y_*\vert \mu_{\phi_1}+\epsilon_{1}\sigma_{\phi_{1}},\mu_{\phi_2}+\epsilon_{2}\sigma_{\phi_{2}},x_*)\big] \\
        -\mathbb{E}_{\epsilon_{1}\sim N(0,I)}\frac{\partial}{\partial\phi_{2}}D_{KL}\big[q(\mu_{\phi_2}+\epsilon_{2}\sigma_{\phi_{2}}\vert \mu_{\phi_1}+\epsilon_{1}\sigma_{\phi_{1}},x_*,y_*)\parallel p(\mu_{\phi_2}+\epsilon_{2}\sigma_{\phi_{2}}\vert \mu_{\phi_1}+\epsilon_{1}\sigma_{\phi_{1}},x_*)\big] 
    \end{split}
    \tag{D.5}
\end{equation}

\begin{equation}\label{eq_14}
    \begin{split}
        \frac{\partial\mathcal{L}}{\partial\theta}=\mathbb{E}_{\epsilon_{1}\sim N(0,I)}\mathbb{E}_{\epsilon_{2}\sim N(0,I)}\frac{\partial}{\partial\theta}\ln\big[p_{\theta}(y_*\vert \mu_{\phi_1}+\epsilon_{1}\sigma_{\phi_{1}},\mu_{\phi_2}+\epsilon_{2}\sigma_{\phi_{2}},x_*)\big] 
    \end{split}
    \tag{D.6}
\end{equation}


\section{Implementation Details in Experiments}
Unless explicitly mentioned, otherwise we make of an one-step amortized transformation as $dim\_lat\mapsto [\mu\_{lat},\ln\sigma\_{lat}]$ to approximate parameters of the posterior in NP models. Especially for DSVNP, the approximate posterior of a local latent variable is learned with the neural network transformation in the approximate posterior $[dim\_lat,dim\_latx,dim\_laty]\mapsto dim\_lat$ and in the piror network $[dim\_lat,dim\_latx]\mapsto dim\_lat$. (For the sake of simplicity, these are not further mentioned in tables of neural structures.) All models are trained with Adam, implemented on Pytorch.

\begin{table*}[t]
\caption{Pointwise Average Negative Log-likelihoods for 2000 realisations. Rows with J consider all data points including the context, while those with P exclude the context points in statistics. (Figures in brackets are variances.)}
\label{sim_results}
\vskip 0.15in
\begin{center}
\begin{small}
\begin{sc}
\begin{tabular}{llllll}
\toprule
Prediction &CNP &NP &AttnNP &DSVNP \\
\midrule
Inter(J) &NaN &-0.958(2e-5) &\textbf{-1.149}\textbf{(8e-6)} &-0.975(2e-5) \\
Inter(P) &-0.802(1e-6) &-0.949(2e-5) &\textbf{-1.141}\textbf{(6e-6)} &-0.970(2e-5) \\
Extra(J) &NaN &8.192(7e1) &8.091(7e2) &\textbf{4.203}\textbf{(9e0)} \\
Extra(P) &\textbf{1.764}\textbf{(1e-1)} &8.573(8e1) &8.172(7e2) &4.303(1e1) \\
\bottomrule
\end{tabular}
\end{sc}
\end{small}
\end{center}
\vskip -0.1in
\end{table*}

\begin{table*}[t]
\caption{Tested Entropies of Logit Probability on Classification Dataset. For rows of MNIST and CIFAR10, the second figures in columns are classification accuracies. Both MC-Dropout and DSVNP are averaged with 100 Monte Carlo samples.}
\label{ood_results}
\vskip 0.15in
\begin{center}
\begin{small}
\begin{sc}
\begin{tabular}{lllllll}
\toprule
   &NN &MC-Dropout &CNP &NP &AttnNP &DSVNP\\
\midrule
MNIST &0.011/0.990 &\textbf{0.009/0.993} &0.019/0.993 &0.010/0.991 &0.012/0.989 &0.027/0.990\\
\midrule
FMNIST &0.385 &0.735 &0.711 &0.434 &0.337 &\textbf{0.956} \\

KMNIST &0.282 &0.438 &0.497  &0.322 &0.294 &\textbf{0.545} \\

Gaussian &0.601 &\textbf{1.623} &1.313 &0.588 &0.611 &0.966 \\

Uniform &0.330 &\textbf{1.739} &0.862  &0.094 &0.220 &0.375 \\

\midrule
CIFAR10 &0.151/0.768 &0.125/0.838 &0.177/0.834 &0.124/0.792 &0.124/0.795 &\textbf{0.081/0.863} \\
\midrule
SVHN &0.402 &0.407 &\textbf{0.459} &0.315 &0.269 &0.326 \\
Rademacher &0.021 &0.062 &0.079  &0.078 &0.010 &\textbf{0.146} \\
Gaussian &0.351 &0.266 &\textbf{0.523} &0.451 &0.349 &0.444 \\
Uniform  &0.334 &0.217 &\textbf{0.499} &0.463 &0.261 &0.374 \\
\bottomrule
\end{tabular}
\end{sc}
\end{small}
\end{center}
\vskip -0.1in
\end{table*}

\subsection{Synthetic Experiments}
For synthetic experiments, all implementations resemble that in Attentive NPs \footnote{https://github.com/deepmind/neural-processes}. And the neural structures for NPs are reported in Table (\ref{Syn_NS}), where dim\_lat is 128. Note that for the amortized transformations in encoders of NP, AttnNP and DSVNP, we use the network to learn the distribution parameters as: $dim\_lat\mapsto [\mu_{lat},{\ln\sigma}_{lat}]$.
In training process, the maximum number of iterations for all (C)NPs is 800k,, and the learning rate is 5e-4. For testing process, in interpolation tasks, the maximum number of context points is 50, while that in extrapolation tasks is 200. Note that the coefficient for KL divergence terms in (C)NPs is set 1 as default, but for DSVNP, we assign more penalty to KL divergence term of local latent variable to avoid overfitting, where the weight is set $\beta_{2}=1000$ for simplicity. Admittedly, more penalty to such term reduces prediction accuracy and some dynamically tuning such parameter would bring some promotion in accuracy.

\begin{table*}[t]
\caption{Neural Network Structure of (C)NP Models for 1-D Stochastic Process. The transformations in the table are in linear form, followed with ReLU activation mostly. And Dropout rate for DNN is defined as 0.5 for all transformation layers in Encoder. As for AttnNP and DSVNP, the encoder network is doubled in the table since there exist some local variable for prediction.}
\label{Syn_NS}
\vskip 0.15in
\begin{center}
\begin{small}
\begin{sc}
\begin{tabular}{lll}
\toprule
NP Models &Encoder &Decoder \\
\midrule
 CNP\&NP  &$[dim\_x,dim\_y]\mapsto 32\mapsto 32\mapsto dim\_lat$ &$[dim\_x,dim\_lat]\mapsto 2*dim\_y$ \\
\midrule
  AttnNP\&DSVNP &$[dim\_x,dim\_y]\mapsto 32\mapsto 32\mapsto dim\_lat$ &$[dim\_x,2*dim\_lat]\mapsto 2*dim\_y$ \\
\bottomrule
\end{tabular}
\end{sc}
\end{small}
\end{center}
\vskip -0.1in
\end{table*}

\subsection{System Identification Experiments}
In the Cart-Pole simulator, the input of the system is the vector of the coordinate/angle and their first order derivative and a random action $[x_c,\theta,x_c^{\prime},\theta^{\prime},a]$, while the output is the transited state in the next time step $[x_c,\theta,x_c^{\prime},\theta^{\prime}]$. The force as the action space ranges between [-10,10] N, where the action is a randomly selected force value to impose in the system. For training dataset from 6 configurations of environments, we sample 100 state transition pairs for each configuration as the maximum context points, and these context points work as identification of a specific configuration. The neural architectures for CNP, NP, AttnNP and DSVNP refer to Table (\ref{Physics_Sim}), and default parameters are listed in $\{dim\_latxy=32,dim\_lat=32,dim\_h=400\}$. All neural network models are trained with the same learning rate 1e-3. The batch size and the maximum number of epochs are 100. For AttnNP, we notice the generalization capability degrades with training process, so early stopping is used. For DSVNP, the weight of regularization is set as $\{\beta_1=1,\beta_2=5\}$, while the KL divergence term weight is fixed as 1 for NP and AttnNP.

\begin{table*}[t]
\caption{Neural Network Structure of (C)NP Models in System Identification Tasks. The transformations in the table are linear, followed with ReLU activation mostly. As for AttnNP and DSVNP, the encoder network is doubled in the table since there exist some local variable for prediction. Here only dot product attention is used in AttnNP.}
\label{Physics_Sim}
\vskip 0.15in
\begin{center}
\begin{small}
\begin{sc}
\begin{tabular}{lll}
\toprule
NP Models &Encoder &Decoder \\
\midrule
        &$[dim\_x,dim\_y]\mapsto \underbrace{dim\_latxy\mapsto dim\_latxy}_{2\;times}$ &$[dim\_x,dim\_lat]\mapsto dim\_h\mapsto dim\_h$ \\
CNP\&NP &$dim\_latxy\mapsto dim\_lat$. &$dim\_h\mapsto 2*dim\_y$ \\
\midrule
               &$[dim\_x,dim\_y]\mapsto \underbrace{dim\_latxy\mapsto dim\_latxy}_{2\;times}$ &$[dim\_x,2*dim\_lat]\mapsto dim\_h\mapsto dim\_h$ \\
 AttnNP\&DSVNP &$dim\_x\mapsto dim\_latx$; & \\
               &$[dim\_latx,dim\_laty]\mapsto dim\_lat$. &$dim\_h\mapsto 2*dim\_y$\\
\bottomrule
\end{tabular}
\end{sc}
\end{small}
\end{center}
\vskip -0.1in
\end{table*}

\subsection{Multi-output Regression Experiments}
SARCOS records inverse dynamics for an anthropomorphic robot arm with seven degree freedoms, and the mission is to predict 7 joint torques with 21-dimensional state space (7 joint positions, 7 joint velocities and 7 accelerations). WQ targets at predicting the relative representation of plant and animal species in Slovenian rivers with some physical and chemical water quality parameters. SCM20D is some supply chain time series dataset for many products, while SCFP records some online click logs for several social issues.

Before data split, standardization over input and output space is operated on dataset, scaling each dimension of dataset in zero mean and unit variance \footnote{https://scikit-learn.org/stable/modules/preprocessing.html}. Such pre-processing is required to ensure the stability of training. Also, we find directly treating the data likelihood term as some Gaussian and optimizing negative log likelihood of Gaussian to learn both mean and variance do harm to the prediction, hence average MSE is selected as the objective. As for the variance estimation for uncertainty, Monte Carlo estimation can be used. For all dataset, we employ the neural structure in Table (\ref{MOR_NS}), and default parameters in Encoder and Decoder are in the list \{$dim\_h=100,dim\_latx=32,dim\_laty=8,dim\_lat=64$\}. The learning rate for Adam is selected as 1e-3, the batch size for all dataset is 100, the maximum number of context points is randomly selected during training, and the maximum epochs in training are up to the scale of dataset and convergence condition. Here the maximum epochs are respectively 300 for SARCOS, 3000 for SCM20D and 5000 for WQ. For the testing process, 30 data points are randomly selected as the context for prediction in each batch.
Also notice that, the hyper-parameters as the weights of KL divergence term are the same in implementation as one without additional modification in this experiments.  

\begin{table*}[t]
\caption{Neural Network Structure of (C)NP Models in Multi-Output Regression Tasks. The transformations in the table are linear, followed with ReLU activation mostly. And Dropout rate for DNN is defined as 0.01 for all transformation layers in Encoder. As for AttnNP and DSVNP, the encoder network is doubled in the table since there exist some local variable for prediction. Here only dot product attention is used in AttnNP.}
\label{MOR_NS}
\vskip 0.15in
\begin{center}
\begin{small}
\begin{sc}
\begin{tabular}{lll}
\toprule
NP Models &Encoder &Decoder \\
\midrule
DNN(MC-Dropout)     &$dim\_x\mapsto \underbrace{dim\_h\mapsto dim\_h}_{2\;times}$ &$dim\_lat\mapsto dim\_h$ \\
        &$dim\_h\mapsto dim\_lat$ &$dim\_h\mapsto dim\_y$ \\
\midrule
        &$dim\_x\mapsto \underbrace{dim\_h\mapsto dim\_h}_{2\;times}\mapsto dim\_latx$ &$[dim\_latx,dim\_lat]\mapsto dim\_h$ \\
CNP\&NP &$dim\_y\mapsto dim\_laty$; & \\
        &$[dim\_latx,dim\_laty]\mapsto dim\_lat$. &$dim\_lat\mapsto dim\_y$ \\
\midrule
               &$dim\_x\mapsto \underbrace{dim\_h\mapsto dim\_h}_{2\;times}\mapsto dim\_latx$ &$[dim\_latx,2*dim\_lat]\mapsto dim\_h$ \\
 AttnNP\&DSVNP &$dim\_y\mapsto dim\_laty$; & \\
               &$[dim\_latx,dim\_laty]\mapsto dim\_lat$. &$dim\_lat\mapsto dim\_y$\\
\bottomrule
\end{tabular}
\end{sc}
\end{small}
\end{center}
\vskip -0.1in
\end{table*}

\subsection{Image Classification and O.O.D. Detection}
The implementations of NP related models and Monte-Carlo Neural Network are quite similar. On MNIST task, the feature extractor for images is taken as LeNet-like structure as [20, 'M', 50, 'F', '500']\footnote{Numbers are dimensions of Out-Channel with kernel size 5, 'F' is flattening operation, and each layer is followed with ReLU activation.}, and the decoder is one-layer transformation. 
On CIFAR10 task, the extractor is parameterized in VGG-style network as [64, 64, 'M', 128, 128, 'M', 256, 256, 256, 256, 'M', 512, 512, 512, 512, 'M', 512, 512, 512, 512, 'M']\footnote{Numbers are dimensions of Out-Channel with kernel size 3 and padding 1 in each layer, followed with BatchNorm and ReLU function, here M means max-polling operation.}, and the decoder is also one-layer transformation from latent variable to label output in softmax form. Other parameters are in the list \{$dim\_latx=32,dim\_laty=64,dim\_lat=64$\} for MNIST and \{$dim\_latx=32,dim\_laty=64,dim\_lat=128$\}.
The labels for both are represented in one-hot encoding way and then further transform to some continuous embedding. Batch size in training is 100 as default, the number of context samples for NP related models is randomly selected no larger than 100 in each batch, while the optimizer Adam is with learning rate $1e^{-3}$ for MNIST task and $5e^{-5}$ for CIFAR10 task. The maximum epochs for both is 100 in both cases, and the size of all source and o.o.d. dataset is 10000. Dropout rates for MC-Dropout in encoder networks are respectively as 0.1 and 0.2 for LeNet-like one and VGG-like one. In the testing process, 100 samples from source dataset are randomly selected as the context points.

\begin{table*}[t]
\caption{Neural Network Structure of (C)NP Models in Image Classification Tasks. The transformations in the table are linear, followed with ReLU activation mostly. And Dropout rate for DNN is defined as 0.5 for all transformation layers in Encoder. As for AttnNP and DSVNP, the encoder network is doubled in the table since there exist some local variable for prediction.}
\label{Image_Classification}
\vskip 0.15in
\begin{center}
\begin{small}
\begin{sc}
\begin{tabular}{lll}
\toprule
NP Models &Encoder &Decoder \\
\midrule
DNN(MC-Dropout)     &$\underbrace{dim\_x\mapsto dim\_h}_{embedding\;net}$ &$dim\_lat\mapsto dim\_y$ \\
        &$dim\_h\mapsto dim\_lat$ \\
\midrule
        &$\underbrace{dim\_x\mapsto dim\_h}_{embedding\;net}\mapsto dim\_latx$ & \\
CNP\&NP &$dim\_y\mapsto dim\_laty$; &$[dim\_latx,dim\_lat]\mapsto dim\_y$ \\
        &$[dim\_latx,dim\_laty]\mapsto dim\_lat$. & \\
\midrule
               &$\underbrace{dim\_x\mapsto dim\_h}_{embedding\;net}\mapsto dim\_latx$ & \\
 AttnNP\&DSVNP &$dim\_y\mapsto dim\_laty$; &$[dim\_latx,2*dim\_lat]\mapsto dim\_y$ \\
               &$[dim\_latx,dim\_laty]\mapsto dim\_lat$. &\\
\bottomrule
\end{tabular}
\end{sc}
\end{small}
\end{center}
\vskip -0.1in
\end{table*}

Before prediction process (estimating predictive entropies on both domain dataset and o.o.d dataset), images on MNIST are normalized in interval $[0,1]$, those on CIFAR10 are standarized as well, and all o.o.d. dataset follow similar way as that on MNIST or CIFAR10. More specifically, Rademacher dataset is generated in the way: place bi-nominal distribution with probability 0.5 over in image shaped tensor and then minus 0.5 to ensure the zero-mean in statistics. Similar operation is taken in uniform cases, while Gaussian o.o.d. dataset is from standard Normal distribution. 
All results are reported in Table (\ref{ood_results}).

\end{document}